\documentclass[letterpaper, 10 pt, conference]{ieeeconf}  

\IEEEoverridecommandlockouts                              

\usepackage{xcolor} 
\usepackage{amsmath}
\usepackage{algorithm} 
\usepackage{algpseudocode} 

\definecolor{darkgreen}{rgb}{0,0.5,0} 
\definecolor{custompurple}{RGB}{128,0,128} 

\definecolor{darkorange}{RGB}{255,140,0}

\newcommand{\darkorangebold}[1]{\textbf{\textcolor{darkorange}{#1}}}
\usepackage{graphicx}
\usepackage{amssymb}
\usepackage{booktabs}
\usepackage{afterpage}
\usepackage{lipsum} 
\usepackage{url}
\usepackage{hyperref}
\usepackage{booktabs}
\usepackage{caption}
\usepackage[acronym,nomain,nonumberlist]{glossaries}
\makeglossaries
\newacronym{NeRF}{NeRF}{Neural Radiance Fields}

\newacronym{3DGS}{3DGS}{3D Gaussian Splatting}
\newacronym{GS}{GS}{Gaussian Splatting}
\newacronym{SDF}{SDF}{Signed Distance Field}

\newacronym{SfM}{SfM}{Structure from Motion}

\newacronym{GPIS}{GPIS}{Gaussian Process Implicit Surface}
\newacronym{GP}{GP}{Gaussian Process}

\title{\LARGE \bf
 Touch-GS: Visual-Tactile Supervised 3D Gaussian Splatting} 

\author{Aiden Swann$^{\ast}{}^{1}$, Matthew Strong$^{\ast}{}^{2}$, Won Kyung Do$^{1}$, Gadiel Sznaier Camps$^{3}$, \\ Mac Schwager$^{3}$, Monroe Kennedy III$^{1,2}$
\thanks{This research was supported by NSF Graduate Research Fellowship No.
DGE-2146755 and NSF Grant No. 2142773, 2220867.}
\thanks{$[\cdot]^{1}$ are with the Department of Mechanical Engineering,
$[\cdot]^{2}$ Department of Computer Science,
        $[\cdot]^{3}$ Department of Aeronautics and Astronautics,
        Stanford University, Stanford CA, USA.
        Emails: \{swann, mastro1, wkdo, gsnaier, schwager, monroek\}@stanford.edu}
\thanks{$^\ast$Both authors contributed equally to this work.}
}

\begin{document}

\maketitle
\thispagestyle{empty}
\pagestyle{empty}

\begin{abstract}
In this work, we propose a novel method to supervise 3D Gaussian Splatting (3DGS) scenes using optical tactile sensors. Optical tactile sensors have become widespread in their use in robotics for manipulation and object representation; however, raw optical tactile sensor data is unsuitable to directly supervise a 3DGS scene. Our representation leverages a Gaussian Process Implicit Surface to implicitly represent the object, combining many touches into a unified representation with uncertainty. We merge this model with a monocular depth estimation network, which is aligned in a two stage process, coarsely aligning with a depth camera and then finely adjusting to match our touch data. For every training image, our method produces a corresponding fused depth and uncertainty map. Utilizing this additional information, we propose a new loss function, variance weighted depth supervised loss, for training the 3DGS scene model. We leverage the DenseTact optical tactile sensor and RealSense RGB-D camera to show that combining touch and vision in this manner leads to quantitatively and qualitatively better results than vision or touch alone in a few-view scene syntheses on opaque as well as on reflective and transparent objects. Please see our project page at \href{http://armlabstanford.github.io/touch-gs}{armlabstanford.github.io/touch-gs}.

\end{abstract}

\section{INTRODUCTION}

Accurate 3D scene and object representations are an essential aspect of robotic interactions with an environment. \gls*{NeRF} \cite{DBLP:journals/corr/abs-2003-08934} have gained prominence as an effective 3D representation. \gls*{NeRF}s have been applied to a number of robotics challenges, including path planning \cite{chen2023catnips} and manipulation \cite{suresh2023neural}. The method of \gls*{3DGS} \cite{3dgs} has recently advanced the field by providing high-quality and high-speed training along with real-time rendering. Precise representations and real-time visual reconstruction are important aspects for robots to interact with their environments. However, in most cases, visual-only information is not sufficient to interact with complex objects. Situations as common as reaching into a cluttered drawer or grasping something in the dark necessitate a sense of touch. 

\begin{figure}[t]
    \centering
    \includegraphics[width=.48\textwidth]{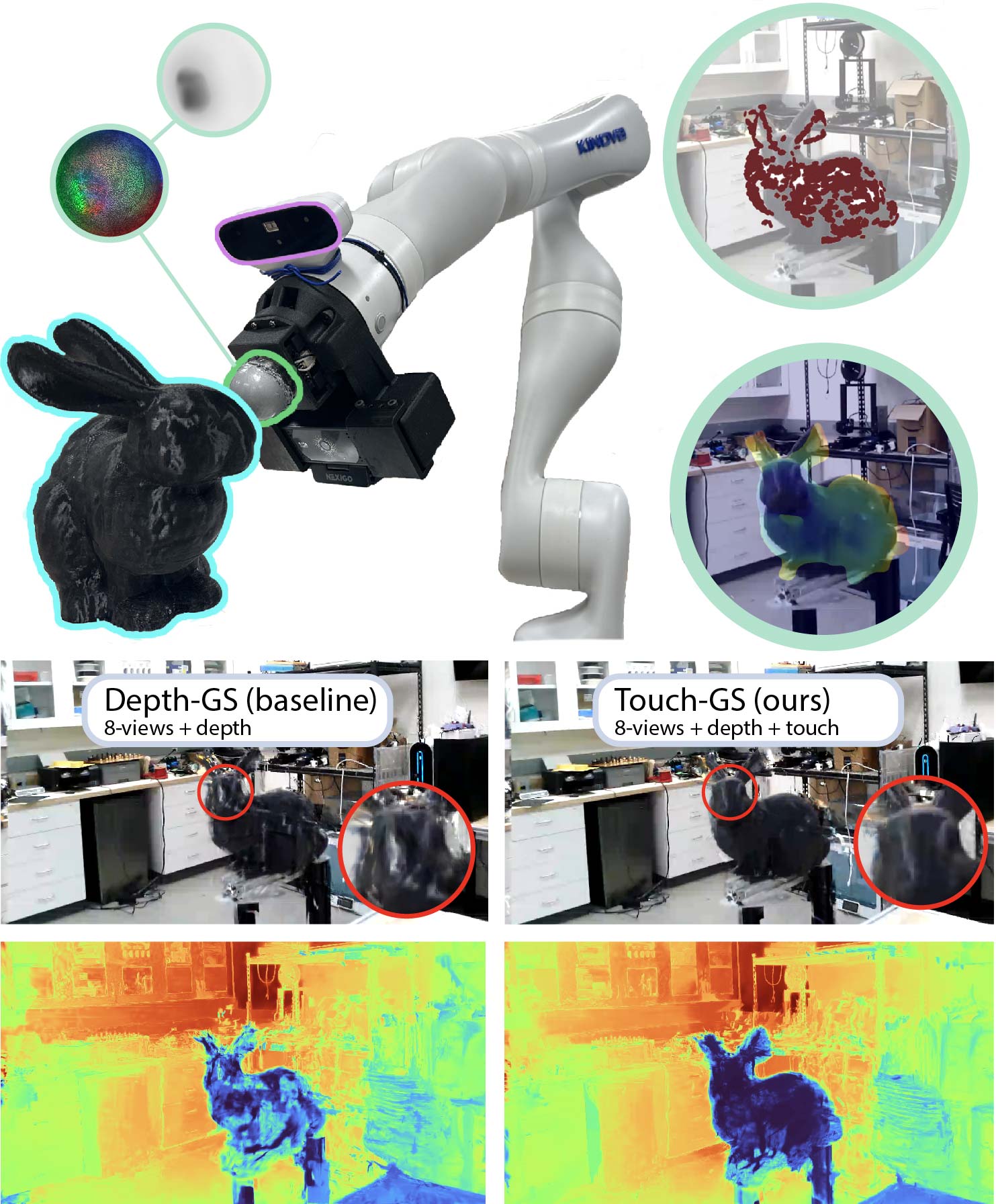}
    \caption{Touch-GS combines monocular depth estimation priors with tactile data-informed implicit surfaces to generate high-quality 3DGS scenes from few training images. Adding touch data significantly enhances 3DGS quality (right) compared with RGB-D alone (left).}
    \label{fig:showcase}
    \vspace{-6mm}
\end{figure}

Recent advances in gel-based optical tactile sensors like DenseTact and GelSight \cite{doDT2.0, donlon2018gelslim} have enabled robots to have a sense of touch. Many works have addressed the use of tactile sensors in robotic applications including topics such as manipulation \cite{do2023inter, qi2023general} and shape reconstruction \cite{comi2023touchsdf, Zhao_Bauza_Adelson_2023}. In this work, we enhance \gls*{3DGS} by using the fusion of visual-tactile data. Just as humans seamlessly integrate both touch and vision to achieve a unified understanding of their environment, robots can leverage this essential integration between touch and vision for interaction with the world.
In a robotic environment, it is not feasible to collect hundreds of images to train a \gls*{3DGS}. In this approach, we utilize both monocular depth
priors and tactile data to generate high-quality few-shot and challenging scenes, such as few-view object-centric scenes, mirrors, and transparent objects.

\textbf{Depth Supervision for \gls*{NeRF}s}. \gls*{NeRF} visual and geometric quality can be improved through depth supervision. 
Several works have utilized \gls*{SfM} keypoints, and monocular depth estimation to train few-view \gls*{NeRF}s \cite{chung2024depthregularized, deng_depth-supervised_2022, roessle2022dense}.
These prior works demonstrate enhanced scene quality, but still require many views to estimate camera poses and sparse keypoints. This was addressed in \cite{fu2023colmapfree}, which does not use COLMAP. 
However, none of these works utilize touch data, which can provide more accurate depth information.

\textbf{Tactile Sensing for Robotic Tasks}. 
Recent research has shown that the high resolution of vision-based tactile sensors allows for rich tactile information and increased sensing capabilities \cite{donlon2018gelslim, doDT2.0}. 
They can be used for object manipulation tasks \cite{xu2023visual,do2023inter}, localization \cite{Zhao_Bauza_Adelson_2023}, shape reconstruction, and even classification of unknown objects \cite{pan2023hand,solano2023embedded}.

\textbf{\gls*{GPIS} Usage for Robot Manipulation}. \gls*{SDF}s are a common implicit approach to represent a 3D object as a 0-level-set of an \gls*{SDF} function. Typically, \gls*{SDF}s are represented as learned functions \cite{park2019deepsdf, comi2023touchsdf}. \gls*{GPIS}s, use a \acrfull*{GP} to represent an \gls*{SDF}  \cite{williams2007gaussian} with applications to both robotic manipulation and object representations \cite{5980395, chen2023sliding}. The advantage of a \gls*{GPIS} is the ability to leverage a continuous object representation derived from discrete touches along with the variance associated with the GP model.

\textbf{Fusing Vision and Touch for Robotics}. In addition to the \gls*{GPIS} capability with tactile information from the sensor, using multi-modal input including vision, depth, or touch has been addressed widely in multiple tasks including creating a shared multimodal space \cite{yang2024binding, fu2024touch}. 
Among the current works, it is important to reconstruct an accurate real-world object mesh to solve the manipulation task \cite{smith2021active, suresh2023neural}. 
Unlike tactile-only or many-view vision-based shape reconstruction \cite{watkins2019multi, smith20203d}, this work aims to reconstruct the object from a small number of RGB images with touch input to estimate an accurate object mesh out of RGB, depth, and touch. 
\subsection{Key Contributions}
We propose Touch-GS, the first approach that combines both tactile and visual data to train \gls*{3DGS}. Our main contributions in Touch-GS are as follows:
\begin{itemize}
    \item We introduce a \acrfull*{GPIS} to synthesize tactile data in a representation suitable for supervising \gls*{3DGS} training. 
    \item We optimally fuse the touch \gls*{GPIS} and monocular depth estimation through Bayesian Inference to create depth and uncertainty images for additional training supervision for the \gls*{3DGS} beyond the RGB images.
    \item We show qualitative and quantitative improvements across a variety of scenes in few-view scene synthesis, including scenes with mirrored and transparent objects.
    \item Our method can supply touch supervision to improve any other Neural Radiance Field representation beyond 3DGS, e.g., Nerfacto \cite{nerfstudio} or the original \gls*{NeRF} \cite{DBLP:journals/corr/abs-2003-08934}. 
\end{itemize}

The remainder of the paper is outlined as follows. 
Section \ref{sec:prelim} first covers mathematical preliminaries and Section \ref{sec:method} introduces our method. Section \ref{sec:results} showcases our results, and finally, Section \ref{sec:conclusions} discusses these results and future avenues of research.

\section{PRELIMINARIES}
\label{sec:prelim}

\subsection{Point-based \gls*{NeRF} and \gls*{3DGS}}

\gls*{3DGS} and other point-based \gls*{NeRF} methods utilize a slightly different representation than standard \gls*{NeRF}s. Rather than represent space as a continuous function where empty space is defined implicitly, in these point based methods, free space as truly empty and image formation is done by blending ordered points with overlapping pixels. We define a ray with camera origin $\mathbf{o}$ and orientation $\mathbf{d}$ as follows:
\begin{equation}
    \label{eqn:ray}
    \mathbf{r}(t) = \mathbf{o} + t\mathbf{d}, \quad t \geq 0,
\end{equation}
The color $C(\mathbf{r})$ and depth $D(\mathbf{r})$ along the ray are computed by blending the set of $\mathcal{N}$ ordered points intersecting the ray:
\begin{equation}
    \hat{C} = \sum_{i \in N} c_i \alpha_i \prod_{j=1}^{i-1} (1 - \alpha_j), \quad \hat{D} = \sum_{i \in N} d_i \alpha_i \prod_{j=1}^{i-1} (1 - \alpha_j),
\end{equation}
where $\alpha_i = 1-\exp(-\sigma_i \delta_i)$, and $d$ is the distance of a particular point along a ray. The following L2 loss between the ground truth (GT) image and volumetric rendered image can be used to optimize the \gls*{NeRF}. 
\begin{equation}
\mathcal{L}_c = \sum_{r \in R} \left\| \hat{C}(r) - C(r) \right\|_2^2
\end{equation}
Regardless of the representation, either volume-based or point-based, these loss functions will work with our method. We can then perform depth supervision in a similar manner to color supervision. 

\subsection{Gaussian Process Implicit Surface}

\begin{figure}[t]
    \centering
    \includegraphics[width=.48\textwidth]{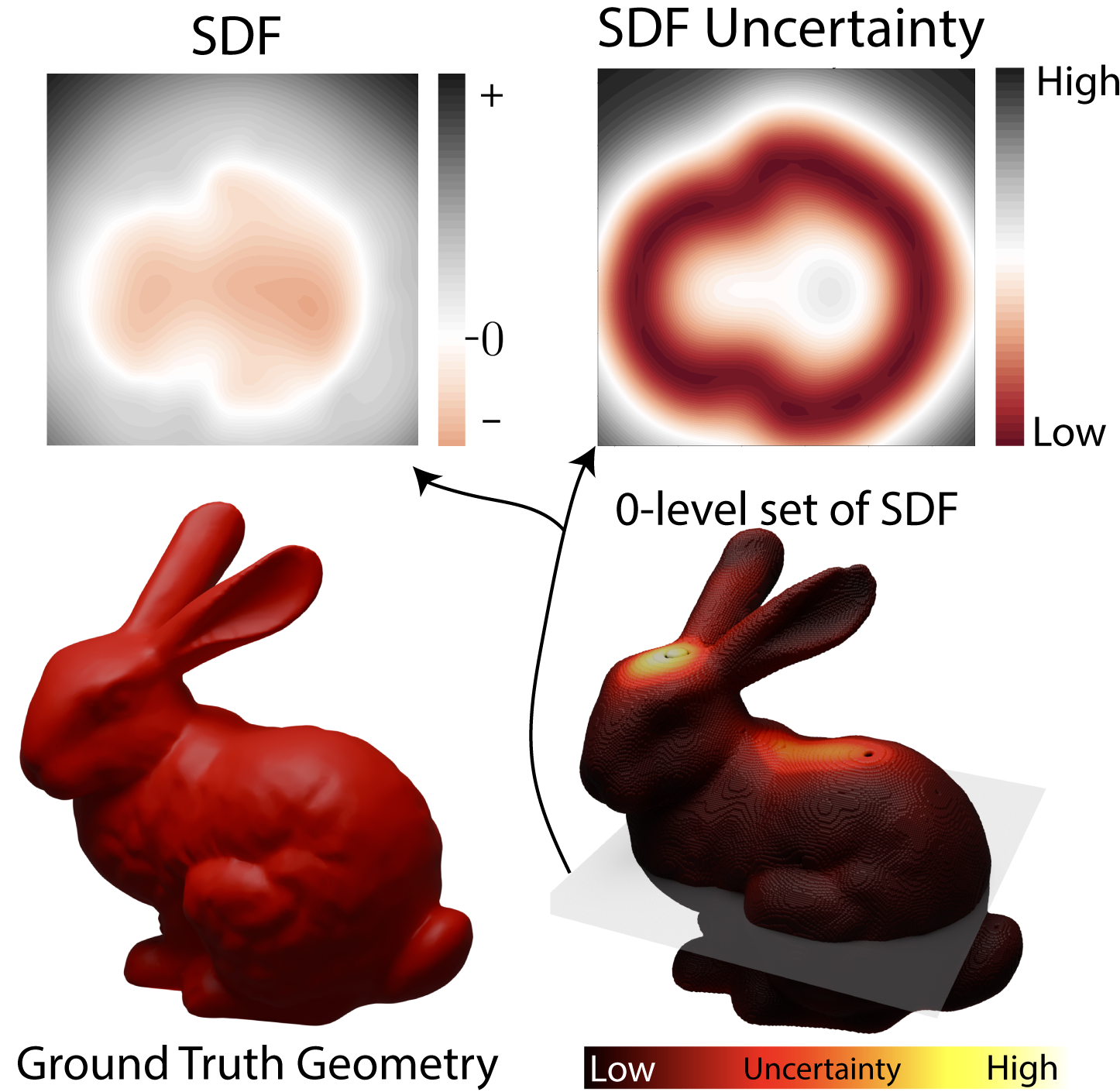}
    \caption{The \gls*{GPIS} is created by finding the 0-level-set of our GP-based SDF. We utilize the uncertainty at the 0-level-set to enhance the accuracy of our model. A z-axis slice of both \gls*{SDF} and uncertainty is shown above the bunnies.}
    \label{fig:GPIS}
    \vspace{-6mm}
\end{figure}

\begin{figure*}[t]
    \centering
    \vspace{.7mm}
    \includegraphics[width=1.0\textwidth]{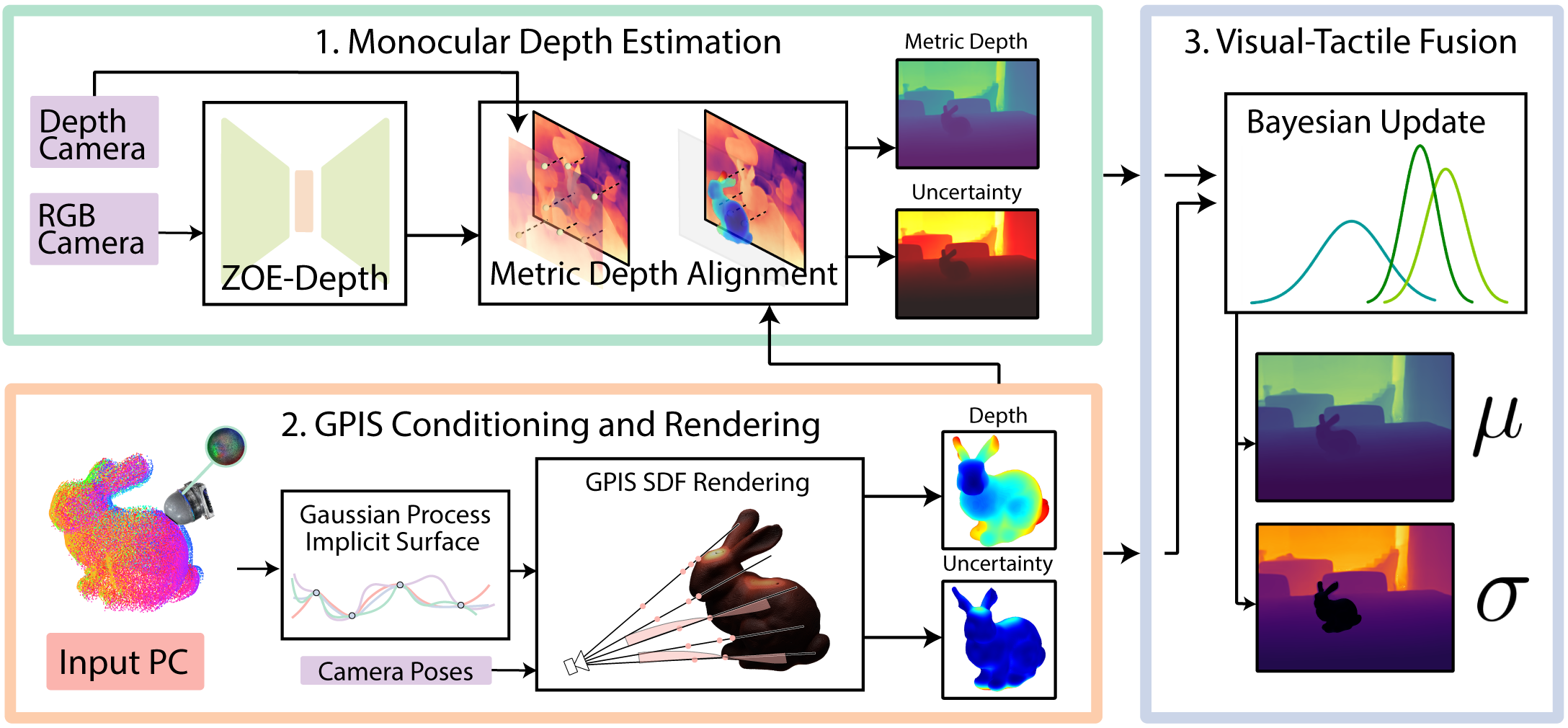}
    \caption{Overview of our method, Touch-GS: 1. We utilize a monocular depth estimation algorithm, which is metrically aligned in a two phase process with RealSense depth and the \gls*{GPIS} output. 2. We condition a GP on the point cloud generated by DenseTact, rendering this into a series of depth and uncertainty images. 3. Monocular depth and tactile information are combined to produce a single set of training images that combine touch and vision.}
    \label{fig:flow-chart}
    \vspace{-4mm}
\end{figure*}

An implicit surface is a surface defined by the 0-level set of a function $f(x)$.
\begin{equation}
   \mathcal{S}_0 \triangleq \{ x \in \mathbb{R}^d \mid f(x) = 0\}.
\end{equation}
In our case, $f(x)$ is a \gls*{SDF}, which maps every point in 3D space to the minimum distance between that point and a given surface. 
We opt for a \gls*{GPIS} because it not only encodes distance information but also captures variance information. GPs are an extremely versatile non-parametric machine learning method. We postulate that our \gls*{SDF} $f(x)$ follows a GP with a given mean and covariance \cite{williams2007gaussian}
\begin{equation}
    f(x) \sim \mathcal{GP}(m(x), k(x, x'))
\end{equation}
The SDF, GP uncertainty, and corresponding 0-level set are shown in Fig. \ref{fig:GPIS}. After making a series of observations, we condition and inference on the GP using a CUDA accelerated library called GPytorch \cite{gardner2018gpytorch}.

\section{METHOD}
\label{sec:method}
Our core method is divided into 3 main modules as shown in Fig. \ref{fig:flow-chart}: 1. Monocular depth estimation on the few-input view RGB data, 2. \gls*{GPIS} conditioning and rendering on the tactile sensor data, and 3. Visual-tactile fusion. These three steps output a single set of depth and variance images. These fused images are then used to train a \gls*{3DGS} model with a slightly modified training and initialization procedure. It is important to note that our approach is modular between different methods of \gls*{NeRF}, point-based or volumetric.

\subsection{\gls*{GPIS} Conditioning and Rendering}
We combine an array of optical tactile sensor measurements, in this case DenseTact, into a single \gls*{GPIS} representation of the object. This fusion of points into a surface is essential for performing supervision on a \gls*{NeRF}. When we touch an object with DenseTact, many areas will be left uncovered, while others will produce conflicting information due primarily to sub-millimeter noise in the manipulator's estimated pose. The \gls*{GPIS} seamlessly combines this information into a 3D estimate of the surface along with uncertainty at every point. 
\subsubsection{\gls*{GPIS} Conditioning}
 In order to produce an accurate \gls*{SDF}, we need to define both the 0-level set and the gradient with respect to it.  We leverage the orientation of the touch to compute the gradient of the surface at each of these points. Following \cite{williams2007gaussian}, we can enforce this during conditioning by generating artificial points both inside and outside the object with negative and positive values respectively:
\begin{equation}
\begin{split}
\mathbf{x}_{in} &= \mathbf{x}_0 - \delta \mathbf{\hat{n}}, \quad f(\mathbf{x}_{in}) = -\delta \\
\mathbf{x}_{out} &= \mathbf{x}_0 + \delta \mathbf{\hat{n}}, \quad f(\mathbf{x}_{out}) = +\delta,
\end{split}
\end{equation}
where $\mathbf{\hat{n}}$ is the normal vector of a given tactile reading which we use as an estimate of the true normal vector $\mathbf{n}$, and $\mathbf{x}_0$ is the set of contact points on the surface of the target object measured by DenseTact, and $\delta$ is a small positive scalar value representing the offset from the surface. 
In addition, we add several points near the center of the object with negative values. These are computed by taking the average position along successive Z-slices of the model.
\begin{equation}
\mathbf{c}_i = \frac{1}{|\mathbf{X}_i|} \sum_{\mathbf{x} \in \mathbf{X}_i} \mathbf{x}, \quad \mathbf{x}_{\text{int},i} = \mathbf{c}_i, \quad f(\mathbf{x}_{\text{int},i}) = -\epsilon
\end{equation}
Where $\mathbf{X}_i$ is the set of points in the $i$-th Z-slice, $\mathbf{c}_i$ is the centroid of the $i$-th Z-slice, and $\epsilon$ is a small positive value used for the \gls*{SDF} value of the interior.
We now condition our \gls*{GPIS} on the following set of points.
\begin{equation}
   \mathcal{X} = \{\mathbf{x}_0, \mathbf{x}_{in}, \mathbf{x}_{out}, \mathbf{x}_{\text{int},i}\}, \quad \mathcal{Y} = \{0, -\delta, +\delta, -\epsilon\}
\end{equation}
While this approximation is not a perfect representation of the true value of the SDF, we find it works well for reasonably shaped objects as shown in Sec. \ref{sec:results}. For our GP covariance function, we use the Matérn kernel with $\nu = 3/2$.
\begin{equation}
    C_{3/2}(d) = \sigma^2 \left(1 + \frac{\sqrt{3}d}{\rho}\right) \exp\left(-\frac{\sqrt{3}d}{\rho}\right)
    \label{eqn:kernel}
\end{equation}
Here, $d$ is the distance between two points and $\rho$ and $\sigma$ are parameters optimizing over during training.

\subsubsection{\gls*{GPIS} Rendering}
\begin{figure}[t]
    \centering
    \includegraphics[width=.48\textwidth]{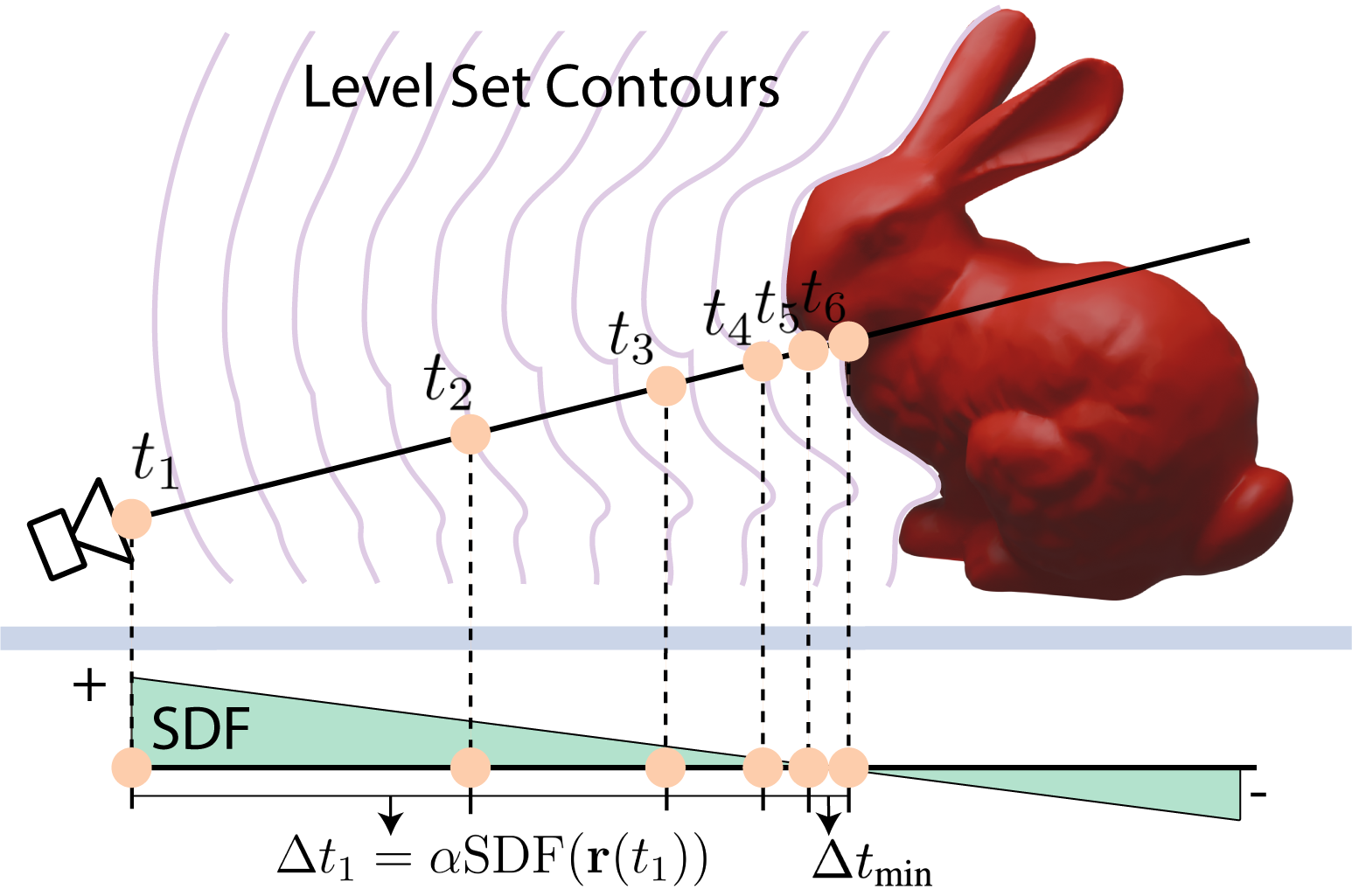}
    \caption{We show our optimized \gls*{SDF} rendering process. $\alpha = .5$, thus each step halves the distance to the surface.}
    \label{fig:GPIS_raymarch}
    \vspace{-5mm}
\end{figure}

We leverage the \gls*{SDF} nature of our conditioned GP to render depth and variance images in the poses of each RGB image. The \gls*{SDF} provides a crucial advantage by indicating the shortest distance to the surface, allowing for an optimized ray-marching that skips empty space. Consider the ray defined in \eqref{eqn:ray}. The ray advances by increments of $\Delta t$ determined by the \gls*{SDF} at the current ray position, until the surface is reached or a maximum number of steps is exceeded, as shown in
\begin{equation}
\label{eqn:deltat}
\Delta t = \max(\alpha\text{SDF}(\mathbf{r}(t)), \Delta t_{\text{min}})
\end{equation}
Where $\Delta t_{\text{min}}$ is the minimum allowable step size and $\alpha$ is a tunable parameter. This variable step size marching is shown in Fig. \ref{fig:GPIS_raymarch}.
\subsubsection{Depth and Variance Image Generation}

\begin{figure}[t]
    \centering
    \includegraphics[width=.48\textwidth]{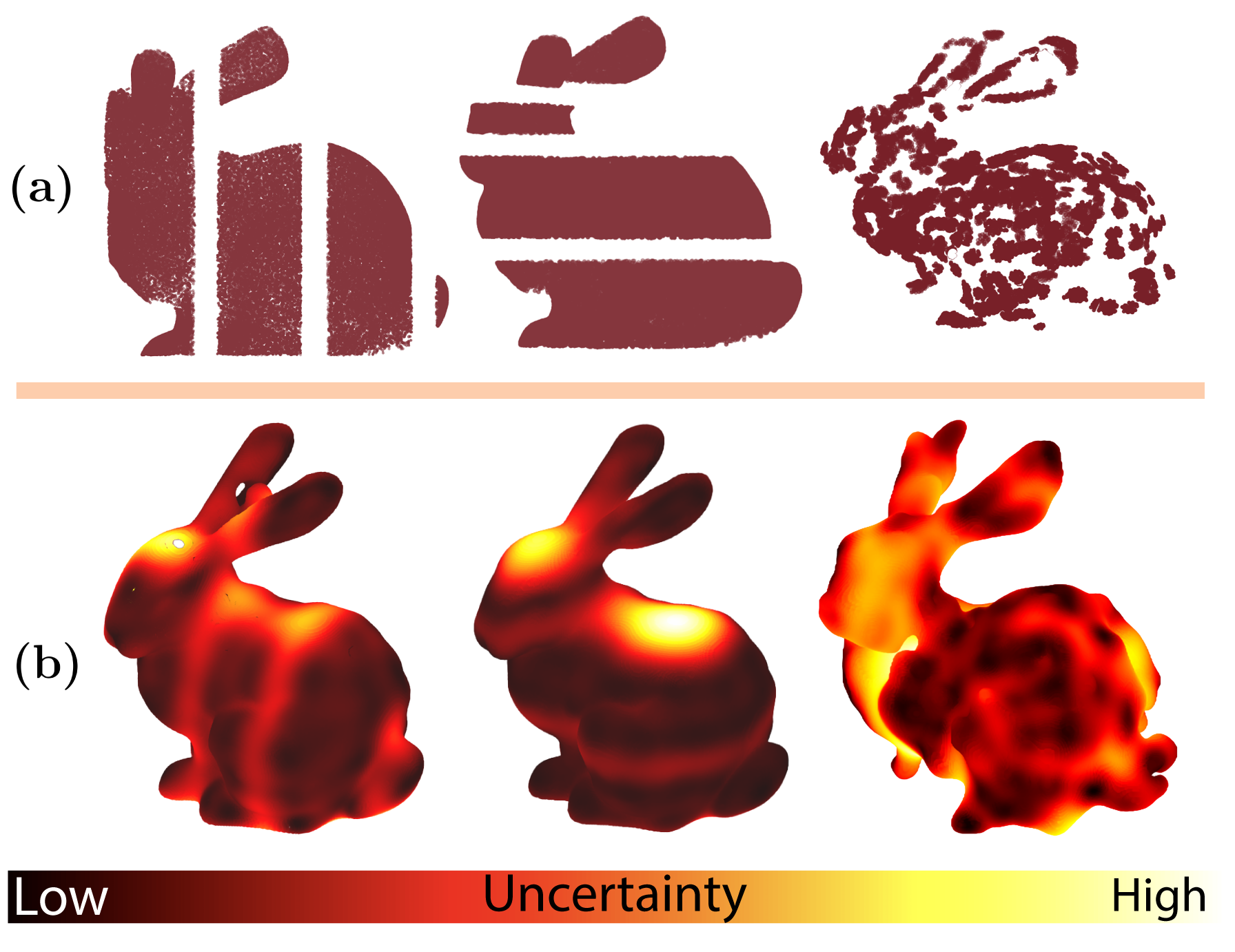}
    \caption{(a) shows the input point clouds and (b) shows the rendered 0-level-set colored by uncertainty. The method is able to fill in the gaps in the point cloud while showing more uncertainty in the interpolated areas. The last column is a real-world dataset.}
    \label{fig:gpis_var}
    \vspace{-6mm}
\end{figure}

To render the \gls*{GPIS} from various viewpoints, we incorporate a standard pinhole camera model which maps from 3D world coordinates into pixel coordinates. 
For each pixel in the RGB image, we compute a corresponding depth value by determining the intersection point of the ray with the object's surface. This ray marching is done by stepping along each $r_{ij}$ in a given camera image following \eqref{eqn:deltat} until the \gls*{SDF} value is within a specified tolerance of $0$. Additional optimizations are done to accelerate the rendering process. Prior to ray marching, we intersect every ray analytically with the minimum volume sphere, which covers a coarse 0-level set. This eliminates rays which will not intersect the surface. The variance image is generated by evaluating the uncertainty in the \gls*{GPIS} at these intersection points. \gls*{GPIS} surfaces along with uncertainties are shown in Fig. \ref{fig:gpis_var} and Fig. \ref{fig:method_images}.

\subsection{Monocular Depth Estimation and Alignment}
While the \gls*{GPIS} representation produces high-quality depth information around the target object, it lacks coverage on the background of the scene. We posit that improving full scene coverage will improve the overall quality of a \gls*{NeRF} reconstruction and geometry of a scene, and even localize the target object in the scene. To supplement this background coverage, we utilize a monocular depth estimation and alignment procedure, as shown in Fig. \ref{fig:flow-chart}, which outputs an estimated absolute scale depth map of the scene. While this closely follows the primary method in \cite{chung2024depthregularized}, our method produces absolute world frame estimates that can be deployed onto a real-world robotic system, where we rely on a robot's kinematics and off-the-shelf depth sensors to align model depth estimations.
\subsubsection{ZoeDepth Estimation and Alignment}

We utilize the state-of-the-art monocular depth estimation network ZoeDepth \cite{bhat2023zoedepth}, but our approach can be applied to any metric or relative monocular depth estimator. Using the raw output of ZoeDepth is not enough; in order to train a \gls*{NeRF}, it is necessary to align the output of ZoeDepth with real-world sparse depth data and touch depth data. To best align the output of ZoeDepth to the scene, we perform a straightforward two-step alignment procedure to construct the most accurate depth map to train a \gls*{NeRF}.

\textit{a)}. \textit{Sparse Depth Scene Alignment}. With an off-the-shelf depth sensor (e.g., the Intel RealSense), we can perform a depth alignment to jointly learn a scale factor and offset to compute an aligned depth map $D_{\text{ZOE}}^*$. Concretely, for each image, we formulate the first alignment stage as a least squares problem shown below:
\begin{equation}
s^*, t^* = \arg\min_{s,t} \sum_{p \in D_{\text{sparse}}} ||D_{\text{sparse}}(p) - D_{\text{ZOE}}(p; s, t)||^2
\end{equation}
where $s^*$ is the scene scale factor, $t^*$ is the offset, and $p$ is a sparse depth keypoint from sparse depth image $D_{\text{sparse}}$.

Our approach also works for \gls*{SfM} algorithms such as COLMAP, which estimate camera poses and outputs a set of sparse keypoints per image. These sparse points are traditionally used as direct supervision or aligning a monocular depth map; however, a successful COLMAP run requires at least 30 images with meaningful visual features to output a list of sparse keypoints. Instead, our method relies on the estimation of camera poses via a robot's kinematics.

\textit{b)}. \textit{Object Alignment}.
While alignment from sparse depth data grounds the overall scale and offset of ZoeDepth, depth data from the \gls*{GPIS} is still useful to further align our vision depth maps. Because the touch data is from one object, we only learn an offset $t_{\text{GPIS}}$ for each depth image. This is implemented as simply reapplying the same linear alignment but constraining the scale to be $1$, and only updating the depth of the object in the ZoeDepth output. This step reshifts the object but does not update meaningful background depths.

\subsubsection{Vision Uncertainty Estimation}
We compute a simple visual uncertainty map as the original depth multiplied by a simple scalar value. Contrary to \cite{roessle2022dense}, our monocular depth estimator does not provide us with a corresponding estimation of depth uncertainty. To this end, we rely on a simple heuristic for uncertainty: farther away depth values should be given more uncertainty, which is reasonable given that the ZoeDepth model is less accurate as distance increases. We also add a simple constant to the result to give touch more priority when fusing the two. Along with a depth map $\mu_{ZOE}$, we now have uncertainty map $\sigma_{ZOE}^2$.

\begin{figure}[t!]
    \centering
    \vspace{1.75mm}
    \includegraphics[width=.48\textwidth]{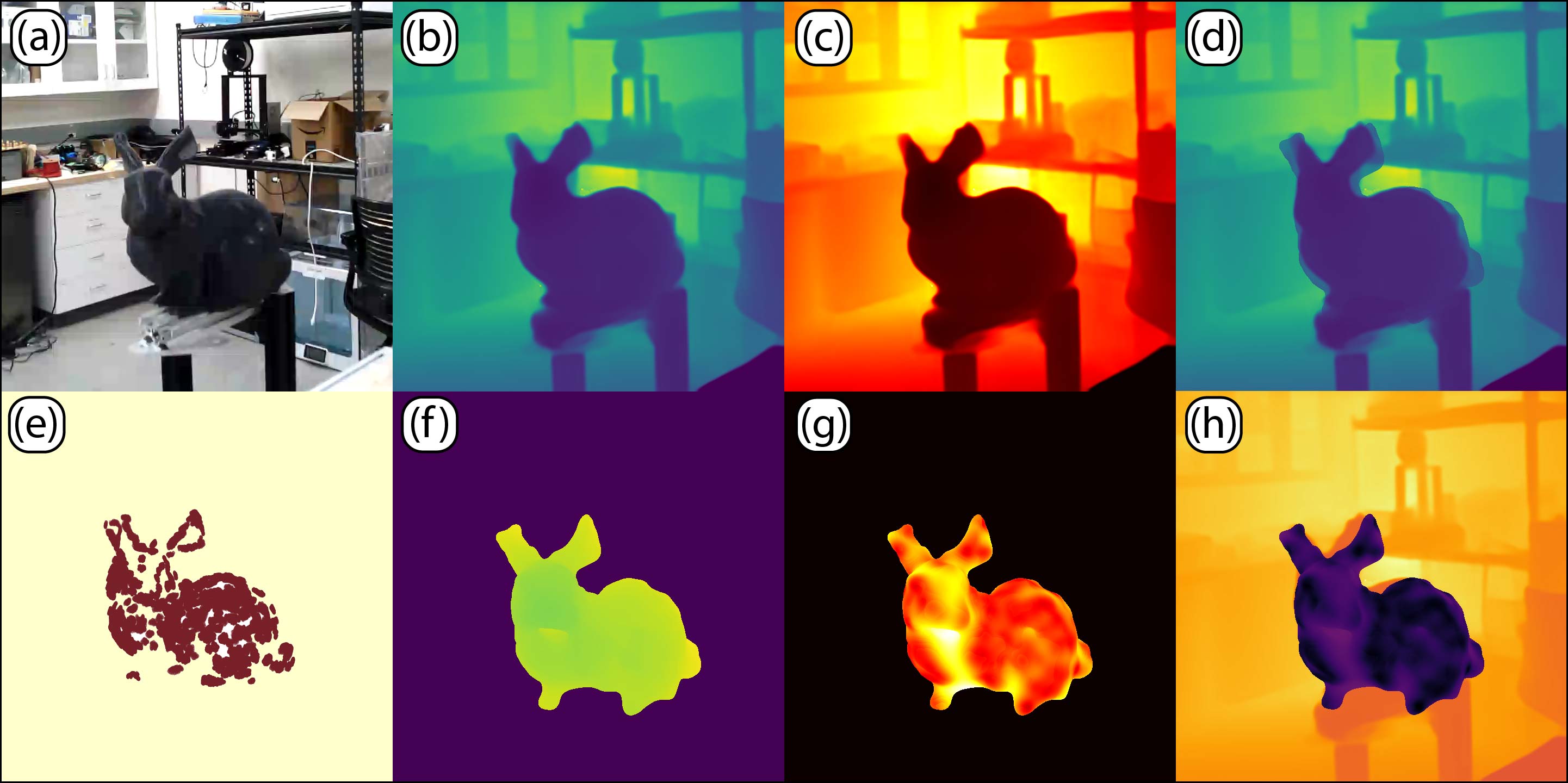}
    \caption{The snapshots of various parts of the method on the real dataset: (a) RGB image, (b) Monocular depth, (c) Monocular uncertainty, (d) Fused touch and monocular depth, (e) DenseTact point cloud, (f) \gls*{GPIS} depth, (g) \gls*{GPIS} uncertainty, (h) Fused touch and monocular uncertainty.}
    \label{fig:method_images}
    \vspace{-5mm}
\end{figure}

\subsection{Depth and Touch Fusion}

    \begin{table*}[t!]
    \vspace{1.5mm}
    \centering
    \begin{tabular}{lccccc}
    \toprule
    Method & PSNR$\uparrow$ & SSIM$\uparrow$ & LPIPS$\downarrow$ & D-MSE$\downarrow$ & D-MSE-O $\downarrow$ \\
    \midrule
    \gls*{3DGS} w/o Depth  & 15.78 & 0.53 & 0.52 & 23.04 & 8.40 \\
    \darkorangebold{Sparse-Depth \cite{deng_depth-supervised_2022}}  & \darkorangebold{18.00} & \darkorangebold{0.54} & \darkorangebold{0.47} & \darkorangebold{13.79} & \darkorangebold{0.92} \\
    \darkorangebold{Dense-Depth \cite{chung2024depthregularized}}  & \darkorangebold{18.21} & \darkorangebold{0.55} & \darkorangebold{0.46} & \darkorangebold{5.77} & \darkorangebold{0.82} \\
    Raw Dense-Depth & 17.37 & 0.55 & 0.45 & 9.38 & 7.48 \\
    Touch-Aligned Vision & 17.99 & 0.56 & 0.46 & 6.39 & 0.20 \\
    Ours Touch Only & 14.32 & 0.54 & 0.57 & 31.07 & 0.46 \\
    Touch w/o Initialization & 17.02 & 0.54 & 0.48 & 21.70 & 0.16 \\
    Ours w/o Initialization, Uncertainty & 18.11 & 0.56 & 0.46 & 6.57 & 0.14 \\
    \textbf{Ours w/o Uncertainty} & \textbf{19.19} & \textbf{0.60} & \textbf{0.42} & \textbf{5.25} & \textbf{0.016} \\
    \textbf{Ours} &  \textbf{19.20} & \textbf{0.60} & \textbf{0.42} & \textbf{5.29} & \textbf{0.016} \\  
    \hline
    \textit{3DGS w/ GT Depth} &  \textit{20.57} & \textit{0.63} & \textit{0.37} & \textit{3.29} & \textit{0.008}\\
    \bottomrule
    \end{tabular}
    \caption{Blender simulated scene ablations with \gls*{3DGS}}    
    \label{tab:results}
    \vspace{-6mm}
    \end{table*}
The output of the parallel pathways of our method are two sets of depth images, both in the same frame, along with corresponding variance information. We combine these two data sources into a single depth and variance supervision pair. This ensures that during training, we weight the touch and monocular depth data relative to their respective uncertainty.
\subsubsection{Bayesian Update}
We treat the problem of fusing the two depth images as a Bayesian update with the dense depth as our prior and the tactile data as our measurement. We assume that the distributions are normally distributed and that each measurement is independent. 
\begin{equation}
\begin{split}
    \sigma_{fuse}^2 &= \left(\frac{1}{\sigma_{ZOE}^2} + \frac{1}{\sigma_{GPIS}^2}\right)^{-1} \\
    \mu_{fuse} &= \sigma_{fuse}^2 \left(\frac{\mu_{ZOE}}{\sigma_{ZOE}^2} + \frac{\mu_{GPIS}}{\sigma_{GPIS}^2}\right)
        \label{eqn:combined_sigma}
\end{split}
\end{equation}
This update rule is applied pixelwise to the outputs of monocular depth estimation and \gls*{GPIS} conditioning / rendering. An example set of input and fused images is shown in Fig. \ref{fig:method_images}.

\subsection{Model Training}

In training our \gls*{NeRF}s, certain depth values should be valued more than others; a depth from touching an object should present a tighter constraint on depth supervision than the estimated depth of an object a few meters away. Additionally, few-input view \gls*{NeRF}s require a good initialization to avoid local minima. Accordingly, we propose a new loss function and novel improvement to the \gls*{3DGS} initialization.

\subsubsection{Uncertainty Weighted Depth Supervision}
With our new variance and depth images, we not only have mean information but also uncertainty; we propose a new cost function, called \textit{uncertainty weighted depth supervision},
\begin{equation}
    \mathcal{L}_d^{(i)} = \alpha^{(i)} \left\| \tilde{d}^{(i)} - d^{(i)} \right\|_2^2 \; \; , \text{where} \; \; \alpha^{(i)} \propto e^{-w\sigma^{(i)}},
\end{equation}

where $\sigma$ is the uncertainty given by the variance image and $w$ is a tunable weight parameter. The total combined loss can be expressed as $\mathcal{L} = \mathcal{L}_{\text{color}} + \lambda \mathcal{L}_d$. This loss function is weighted based on the uncertainty of our \gls*{GPIS} mean. We also add an optional depth loss weight decay shown as $\lambda_{i+1} = \beta \lambda_{i}$, where $\beta$ is a value between $0$ and $1$. We recommend this to get a \gls*{NeRF} out of local minima if the depth weight is too high.

\subsubsection{Model Initialization}

We propose a modified point initialization performed in \gls*{3DGS}. 3D point initialization becomes a much higher influence on training in the few-input view case, and as such, requires an accurate initialization to guide the GS into a geometrically precise and photorealistic scene. We do not have COLMAP data to initalize the \gls*{3DGS} and cannot rely on a potentially noisy depth sensor for initialization, especially in environments where vision is not sufficient, such as reflective surfaces. Thus, for each image $I_k$ in our few input view list, we backproject the depth of each \gls*{GPIS} depth image $\mu_{GPIS, k}$ into the world frame with the robot camera poses, and use the combined point cloud to initialize a \gls*{3DGS}. We also initialize on only touch to highlight its usefulness for an entire scene.
Finally, we observe that in the case of \gls*{3DGS}, convergence is reached quickly, and thus train all \gls*{3DGS} models in our ablations and real-world experiments for 15000 steps for a fair comparison.

\begin{table}[ht]
        \vspace{-1mm}
    \begin{center}
    \begin{tabular}{lccccc}
    \toprule
    Method & PSNR$\uparrow$ & SSIM$\uparrow$ & LPIPS$\downarrow$\hspace{-2mm} & D-MSE-O$\downarrow$ \\
    \midrule
    Nerfacto  & 12.57 & 0.310 & 0.77 & 0.85 \\
    DS-NeRF & 17.16 & 0.58 & 0.46 &  99.17 \\
    Dense Depth  & 18.06 & 0.56 & 0.46 & 1.68 \\
    \textbf{Our Method}  & \textbf{18.37} & \textbf{0.56} & \textbf{0.44} & \textbf{0.60} \\
    \hline
    \end{tabular}
    \caption{Nerfacto Results}
    \label{tab:nerfactoresults}
    \end{center}
        \vspace{-3mm}
\end{table}

\begin{table}[ht]
    \vspace{-2mm}
\begin{center}
\begin{tabular}{lcccccc}
\toprule
Object&Method & PSNR$\uparrow$ & SSIM$\uparrow$ & LPIPS$\downarrow$\hspace{-2mm} \\
\midrule
&\gls*{3DGS}  & 10.30 & 0.41 & 0.61 \\
Real-world&Dense-Depth  & 11.40 & 0.45 & 0.55 \\
Bunny&No Uncertainty  & 11.71 & 0.47 & 0.53 \\
&\textbf{Our Method}  & \textbf{11.75} & \textbf{0.47} & \textbf{0.52} \\
\hline
&\gls*{3DGS}  & 15.39 & 0.65 & 0.42 \\
Mirror&Dense-Depth  & \textbf{15.75} & \textbf{0.66} & \textbf{0.42} \\
&\textbf{Our Method}  & 15.51 & \textbf{0.66} & \textbf{0.42} \\
\hline
&\gls*{3DGS}  & 14.43 & 0.55 & 0.42 \\
Prism&Dense-Depth  & \textbf{14.85} & \textbf{0.57} & \textbf{0.41} \\
&\textbf{Our Method}  & 14.71 & \textbf{0.57} & \textbf{0.41} \\
\hline
\end{tabular}
\caption{Real-world experiment with multiple objects}
\label{tab:realworldresults}
\end{center}
    \vspace{-6mm}
\end{table}

\section{RESULTS}

\label{sec:results}
\subsection{Experimental Setup}

Touch-GS is implemented in both simulated and real-world experimental setups. The simulated evaluation assesses the performance of various ablations in Blender. We obtain our camera poses from Blender and use a point cloud sampled from the GT geometry. To generate sparse points for alignment, we take a small percent (under 1\%) of the GT depth data and perturb it with noise that quadratically increases with distance. We train our scene on 5 equally spaced views around the bunny.

The real-world evaluation assesses the use of Touch-GS and uncertainty in real-world conditions. Touch-GS requires accurate poses for the camera, depth camera, and touch sensor. 
We utilized the RGB and depth cameras installed in the Kinova\textsuperscript{\texttrademark} Gen 3 and attached the DenseTact 2.0 \cite{doDT2.0} at the end-effector of the robot arm as shown in Fig. \ref{fig:showcase}. We collect between 8-151 RGB-D images and between 150-500 touches per object. 
From the reconstructed depth image of the sensor, we determined the contact point by thresholding the estimated point cloud. The final touch data includes between 5,000 to 20,000 points per touch.

\begin{figure*}[t]
    \centering
    \includegraphics[width=.87\textwidth]{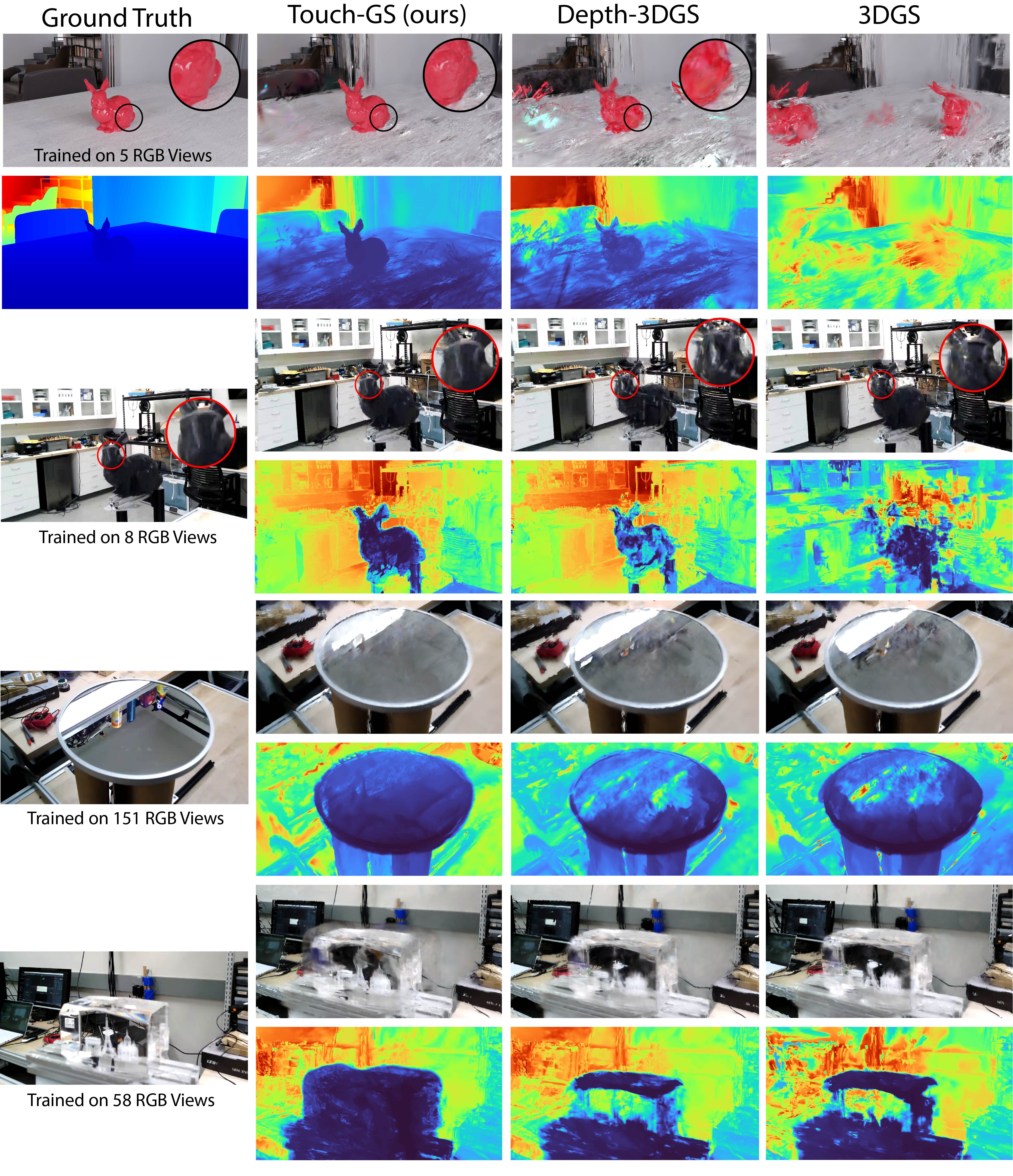}
    \caption{Here we show comparisons of our method to the baselines and ground truth across 1 synthetic and 3 real scenes.}
    \label{fig:results}
    \vspace{-6mm}
\end{figure*}
\subsection{Simulation Evaluation}

We perform an extensive ablation of our method, which includes ablations with GS and a method called Nerfacto, which is the default \gls*{NeRF} method used in Nerfstudio \cite{nerfstudio}.
Our first set of ablations builds up our method from 3DGS.
Our second set of ablations performs an analysis of our method on Nerfacto, which is Nerfstudio's default, volumetric method. Our ablations in GS are listed in Table \ref{tab:results} for brevity. We reimplement \cite{deng_depth-supervised_2022} and \cite{chung2024depthregularized} to the best of our ability; the former method's code was not written for 3DGS, and the latter method's code has not been released. We also tune both methods' depth loss weight to best assess their performance. Finally, we train 3DGS with ground truth depth provided by Blender (and initializing from ground truth bunny points), providing a strong upper bound for the quality of a model trained with perfect depth.
Each method is run 10 times and we report the mean of the standard \gls*{NeRF} metrics (PSNR, SSIM, LPIPs) and ground truth depth and object depth mean square error (MSE), which we call D-MSE and D-MSE-O respectively. 

In Table \ref{tab:results}, 3DGS performs the worst (besides D-MSE), as a lack of depth leads to unrecognizable depths and the inability to localize the bunny. Using sparse depth alone significantly improves visual and depth metrics; indicating how even some form of depth priors improves \gls*{NeRF} generalization. \cite{chung2024depthregularized} outperforms sparse depth in all metrics, most notably the scene ground truth depth MSE, shown in Fig. \ref{fig:results}. Aligning vision to touch leads to slightly worse depth loss but better object depth loss.

The need for alignment is shown in Raw Dense-Depth, where the \gls*{NeRF} performs better than pure GS, but the lack of scale and offset is most noticeable in the poor object ground truth depth MSE. Using only touch improves visual quality without initialization and has an even higher D-MSE-O than when \gls*{GPIS} initialization is performed for just touch. This is due to the model correctly learning the 3D geometry of the bunny but being unable to localize it without background depth. 
When we fuse both vision and touch without initialization, the visual quality is comparable to \cite{chung2024depthregularized} but with a higher depth loss and lower object ground truth depth loss. Once the \gls*{GPIS} point initialization is included, the visual and geometric quality improves dramatically. This is seen clearly in the first column of Fig. \ref{fig:results} in our method, where the bunny geometry is extremely sharp. It is apparent that touch and vision complement each other: vision improves the background which localizes a touched object, and touch improves the background rendering. In Table \ref{tab:nerfactoresults}, our method is the best in all metrics for Nerfacto, demonstrating the value of touch.
\subsection{Real world Evaluation}
For the real-world evaluation, we leave all tunable parameters of the \gls*{GPIS} fixed with the exception of $\rho$ in \eqref{eqn:kernel}, which is adjusted due to the differing length scales of our four scenes. In the vision depth and uncertainty maps, we only update the constant added to vision uncertainty, and we also filter out differences between ZoeDepth and the \gls*{GPIS} larger than 3 meters when we compute the touch-aligned vision map. In training the \gls*{NeRF}, we update the depth and/or uncertainty weight for each scene.  We report mean PSNR, SSIM, and LPIPs with 3 trials for each method. 

In the real-world bunny example, we train a \gls*{NeRF} on 8 input views and test on 40. Without any depth priors, 3DGS struggles, failing to properly render the ears of the bunny and background. The baseline from \cite{chung2024depthregularized} outperforms 3DGS in all visual qualities with noticeably better geometries; however, it fails to completely render the bunny. Our method with uncertainty is able to both render the background and bunny with clear geometries; rendering the tail of the bunny which was seen in touch and never predicted by vision.

\begin{table}[ht]
    \vspace{-2mm}
\begin{center}
\begin{tabular}{lcccc}
\toprule
Method & Chamfer $\downarrow$ & Hausdorff$ \downarrow$ \hspace{-2mm} \\
\midrule
3DGS  & 0.037 & 0.12 \\
Dense-Depth  & 0.027 & 0.14 \\
\textbf{Our Method}  & \textbf{0.023} & \textbf{0.094}\\
\hline
\end{tabular}
\caption{Real-world Bunny Object  Geometric Reconstruction Quality.}
\label{tab:reconstructionresults}
\end{center}
    \vspace{-5mm}
\end{table}

We report the average shape dissimilarity (Chamfer distance) and maximal shape dissimilarity (Hausdorff distance) between the sensed real-world bunny and a ground truth bunny mesh in Table \ref{tab:reconstructionresults}. We compute a transformation, which we refine with Chamfer distance, to best align each method's point cloud to ground truth for a most fair comparison. While Depth-3DGS has better overall shape reconstruction than 3DGS, it suffers from larger outliers, which is due to ZoeDepth incorrectly localizing the bunny. Across all metrics, our approach significantly outperforms 3DGS and Depth-3DGS, highlighting the importance of touch.

The mirror and prism examples serve as a challenging scene for vision. Even with many input views (151 for the mirror and 58 for the prism), the rendered depths from both 3DGS and vision-depth-supervised 3DGS are full of holes. The features shown in the mirror present a challenge for 3DGS and depth GS, which view the reflected features as objects with depth. In the other methods, the prism is transparent. Our method renders a flat, geometry respecting, surfaces on the mirror and prism, and while the visual metrics of \cite{chung2024depthregularized} are better than our method, it is because the high depth weights given to touch make it harder to render the reflections and see-through objects, which we plan to address in future work. But the accurate geometries of these objects in our method present an important first step for robotic manipulators to operate in these environments.

\section{CONCLUSION}
\label{sec:conclusions}
In this work, we have shown how Touch-GS fuses visual-tactile data to produce 3DGS scenes. We demonstrate qualitative and quantitative improvements over our baselines of standard and depth-supervised 3DGS. Robotic systems of the future need to be able to fuse touch and vision to interact with their environment. Our work addresses the fundamental balance between coarse RGB data fused with fine tactile data. In the future, we hope to expand this representation to be dynamic, including both deformability and frictional properties of the object in question to create a true digital twin. 





\bibliographystyle{IEEEtran} 
\bibliography{ms} 

\end{document}